\documentclass[twoside,11pt]{article}

\usepackage{jmlr2e}
\usepackage{enumitem}
\usepackage[utf8]{inputenc}
\usepackage{listings}
\usepackage{xcolor}
\usepackage{subfigure}
\usepackage{bm}
\usepackage{wrapfig}

\usepackage{xspace}
\newcommand{\sys}{\texttt{OpenBox}\xspace}

\newcommand{\presec}{\vspace{-0.75em}}
\newcommand{\postsec}{\vspace{-0.25em}}

\usepackage{lastpage}
\jmlrheading{25}{2024}{1-\pageref{LastPage}}{4/23}{5/24}{23-0537}{Huaijun Jiang, Yu Shen, Yang Li, Beicheng Xu, Sixian Du, Wentao Zhang, Ce Zhang and Bin Cui}

\ShortHeadings{OpenBox: A Python Toolkit for Generalized Black-box Optimization}{Jiang, Shen, Li, Xu, Du, Zhang, Zhang and Cui}
\firstpageno{1}

\begin{document}

\title{OpenBox: A Python Toolkit for Generalized Black-box Optimization}

\author{\name Huaijun Jiang\boldsymbol{$^{1*}$} \email  jianghuaijun@pku.edu.cn \\
    \name Yu Shen\boldsymbol{$^{1*}$} \email shenyu@pku.edu.cn \\
    \name Yang Li\boldsymbol{$^{2*}$} \email thomasyngli@tencent.com \\
    \name Beicheng Xu\boldsymbol{$^{1}$} \email  beichengxu@stu.pku.edu.cn \\
    \name Sixian Du\boldsymbol{$^{1}$} \email  dusixian@stu.pku.edu.cn \\
    \name Wentao Zhang\boldsymbol{$^{1}$} \email wentao.zhang@pku.edu.cn \\
    \name Ce Zhang\boldsymbol{$^{3}$} \email ce.zhang@ethz.ch \\
    \name Bin Cui\boldsymbol{$^{1}$} \email bin.cui@pku.edu.cn \\
    \addr $^{1}$ Key Lab of High Confidence Software Technologies (MOE), School of CS, Peking University, China \\
    \addr $^{2}$ Department of Data Platform, TEG, Tencent Inc., China \\
    \addr $^{3}$ Department of Computer Science, ETH Z{\"u}rich, Switzerland \\
    \addr $^{*}$ Equal contribution. \\
    \vspace{-2.5em}
}

\editor{Zeyi Wen}

\maketitle

\begin{abstract}

Black-box optimization (BBO) has a broad range of applications, including automatic machine learning, experimental design, and database knob tuning. 
However, users still face challenges when applying BBO methods to their problems at hand with existing software packages in terms of applicability, performance, and efficiency.
This paper presents \sys, an open-source BBO toolkit with improved usability. 
It implements user-friendly interfaces and visualization for users to define and manage their tasks. 
The modular design behind \sys facilitates its flexible deployment in existing systems. 
Experimental results demonstrate the effectiveness and efficiency of \sys over existing systems.
The source code of \sys is available at \url{https://github.com/PKU-DAIR/open-box}. 

\end{abstract}


\presec
\section{Introduction}
\postsec

Black-box optimization (BBO)~\citep{munoz2015algorithm} deals with optimizing an objective function under a limited budget for function evaluation. 
However, since the evaluation of objective function is usually expensive, the goal of BBO is to find a configuration that approaches the optimal configuration as soon as possible. 
Recently, generalized BBO has attracted great attraction in various areas~\citep{foster2019variational,sun2022black,sun2022bbtv2,meng2023survey,huang2023survey}, which requires more functionalities than traditional BBO tasks.
Specifically, traditional BBO refers to BBO with integer or float input parameters and with only a single objective, while there can be various input types (e.g., ordinal and categorical), multiple objectives, and constraints in generalized BBO.
Though many software and platforms have been developed for traditional 
BBO~\citep{hutter2011sequential,bergstra2011algorithms,golovin2017google,akiba2019optuna,balandat2020botorch,liu2022zoopt,lindauer2022smac3}, 
so far, there is no platform specifically designed to target generalized BBO. 
Existing platforms suffer from the following limitations when applied to generalized BBO scenarios: 
1) Restricted application scope.
Many existing BBO platforms cannot support multiple objectives and constraints, and they cannot support categorical parameters.  
2) Unstable performance across problems.
Many platforms only implement one or very few BBO algorithms. 
According to the ``no free lunch'' theorem~\citep{ho2001simple}, this would inevitably lead to unstable performance when applied to various problems. 
3) Limited scalability and efficiency. Most platforms execute optimization in a sequential manner, which is inefficient and unscalable. 

This paper proposes \sys to address the above limitations simultaneously. 
\sys has the following crucial features: 
1) \textbf{Generality.} 
\sys hosts most state-of-the-art optimization algorithms and is capable of handling BBO tasks with different requirements. 
2) \textbf{Ease of use.} 
\sys provides user-friendly interfaces, visualization, and automatic decisions on algorithms. 
Users can conveniently define their tasks via either wrapped services or ask-and-tell interfaces. 
3) \textbf{State-of-the-art performance. }
Extensive experiments show that \sys outperforms the existing systems on a wide range of BBO tasks.

So far, \sys has helped researchers solve various realistic BBO problems like database tuning~\citep{kanellis2022llamatune,zhang2022towards} and traffic simulation~\citep{liang2022cblab}. 
It has also performed as the core part of the open-source graph learning system SGL~\citep{zhang2022pasca} and database tuning system DBTune~\citep{zhang2022facilitating}.
Besides the academic usages, \sys won the first place in CIKM 2021 AnalytiCup Track 2~\citep{jiang2021automated}, and parts of the toolkit have been successfully deployed in corporations like Tencent~\citep{li2023towards} and ByteDance~\citep{shen2023rover}.

\presec

\begin{figure*}[h]
    \centering  
    \includegraphics[width=0.95\linewidth]{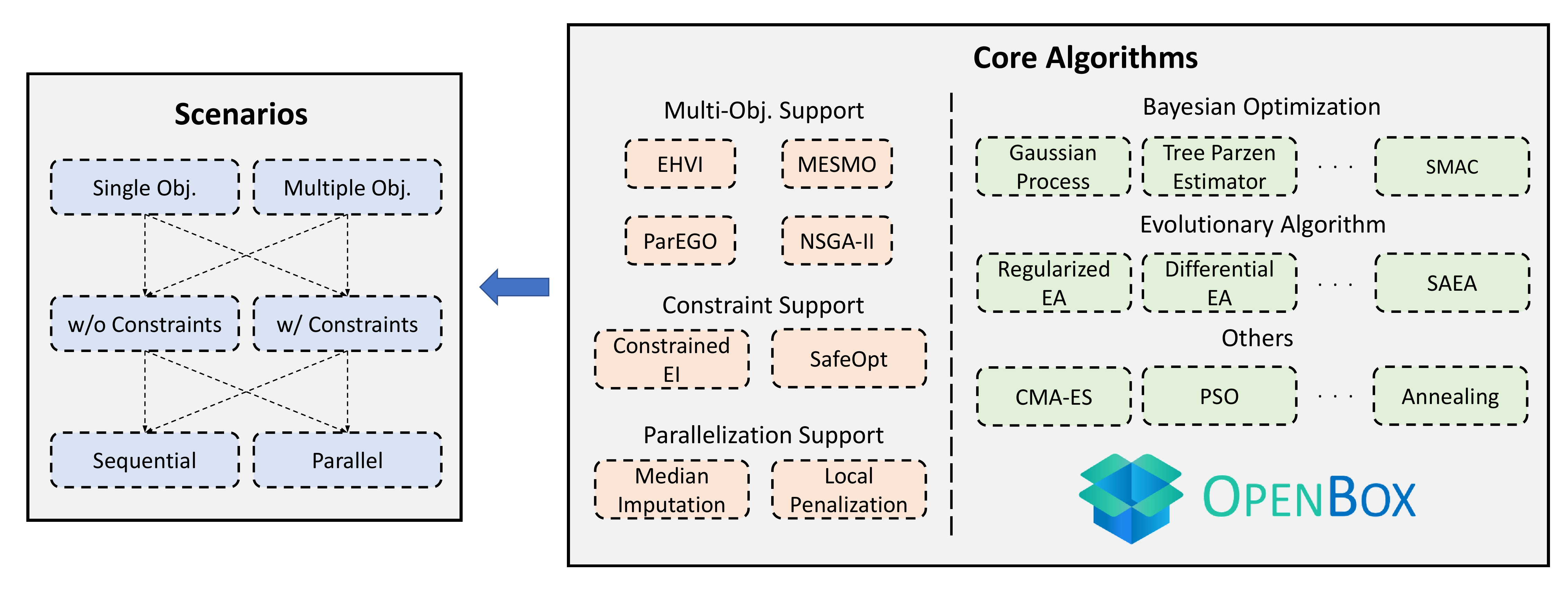}
    \vspace{-1em}
    \caption{Simplified overview of supported scenarios and built-in algorithms in \sys.}
    \label{fig:components}
    \vspace{-1em}
\end{figure*}

\section{Functionality Overview}
\postsec

\label{sec:func}

\sys implements a wide range of state-of-the-art algorithms for various BBO problems (Figure~\ref{fig:components}). 
Concretely, \sys is capable of handling BBO problems with different numbers of objectives~\citep{belakaria2019max,belakaria2020uncertainty} and constraints~\citep{sui2015safe}, and the optimization procedure can either be sequential or parallel~\citep{gonzalez2016batch}.
To ensure stable performance in different scenarios, \sys implements variants of Bayesian optimization~\citep{hutter2011sequential,bergstra2011algorithms,snoek2012practical,hou2022dimensionality}, evolutionary algorithms~\citep{storn1997differential,deb2002fast,jin2011surrogate}, and other competitive methods like CMA-ES~\citep{hansen2003reducing}.

While it may be challenging for users to choose the proper algorithm for different scenarios, \sys also provides automatic decisions on algorithms and settings according to the characteristics of the incoming task,
including the search space, number of objectives, and number of constraints.
The decisions are based on experimental analysis~\citep{eggensperger2013towards} or practical experience.
For example, if there are over ten parameters in the input space, or the number of trials exceeds 300, we choose Probabilistic Random Forest (PRF)~\citep{hutter2011sequential} instead of Gaussian Process (GP) as the surrogate to avoid incompatibility or high computational complexity in Bayesian optimization.

\presec
\section{System Design}
\postsec

\begin{figure*}[ht]
    \centering  
    \includegraphics[width=0.99\linewidth]{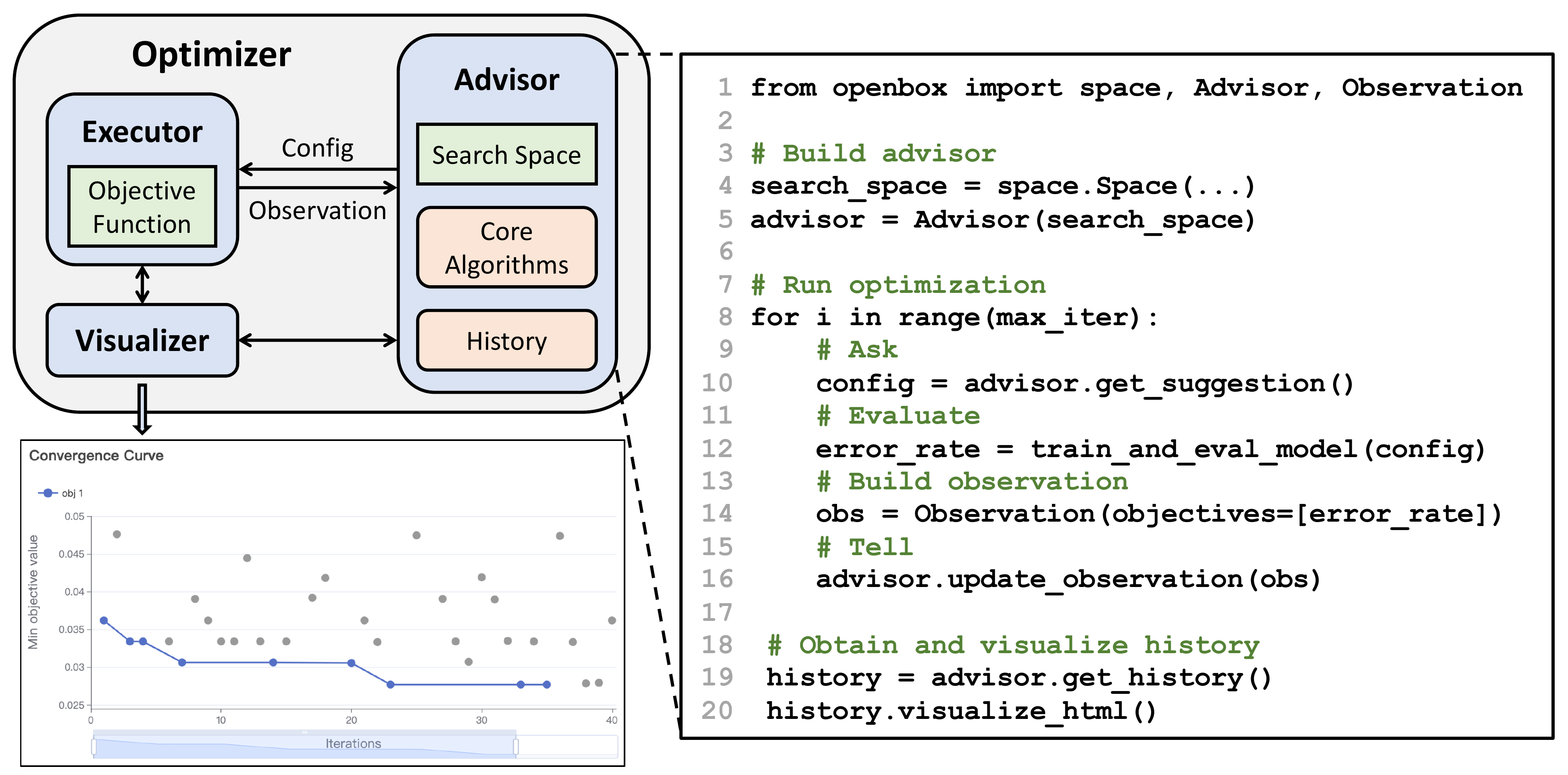}
    \caption{System overview of \sys, including the system architecture (upper left), an example of the ask-and-tell interface using Advisor (right), and an example of visualization interfaces (bottom left).}
    \label{fig:system_design}
\end{figure*}

In this section, we will introduce the system design of \sys.
As shown in the upper left corner of Figure~\ref{fig:system_design}, the core of the optimization framework is \texttt{Optimizer}, which takes an objective function and a search space as inputs, executes the optimization process, and outputs the final results. 
\texttt{Optimizer} consists of three main components: \texttt{Advisor}, \texttt{Executor}, and \texttt{Visualizer}. 
\texttt{Advisor} implements the optimization algorithms described in Section~\ref{sec:func} and suggests new configurations based on the provided search space and stored optimization history. 
\texttt{Executor} then runs the objective function on the recommended configurations and returns the results, i.e., observations. 
To monitor the optimization process, \texttt{Visualizer} provides comprehensive visualization APIs for users, including the convergence curve, parameter importance analysis based on SHAP~\citep{lundberg2017unified}, etc. 
For multi-objective problems, Pareto front and Hypervolume charts are also available. 
An example of the convergence curve is provided on the bottom left of Figure~\ref{fig:system_design}, where the best-observed configurations during optimization are connected by a blue line.

For ease of use, \sys also provides the ask-and-tell interface using \texttt{Advisor}. 
The example code is shown on the right side of Figure~\ref{fig:system_design}. 
Users first build an \texttt{Advisor} based on the search space (Line 4-5). 
Next, users run optimization by interacting with the \texttt{Advisor} (Line 8-16).
Concretely, users get a configuration suggestion from \texttt{Advisor} (Line 10) and evaluate the performance of the configuration by themselves (Line 12). 
Users then build an \texttt{Observation} that contains the evaluation result (Line 14) and update the \texttt{Observation} to \texttt{Advisor} (Line 16). 
Finally, an HTML page for visualization is generated based on the optimization history obtained from \texttt{Advisor} (Line 19-20). 
This ask-and-tell interface is compatible with various types of optimization algorithms and enables users to make flexible modifications to the optimization process.

\presec
\section{Benchmark}
\postsec

\begin{figure*}[tb]
\centering
\subfigure[CONSTR (2 objectives, 2 constraints)]{
    \scalebox{0.48}{
        \includegraphics[width=0.97\linewidth]{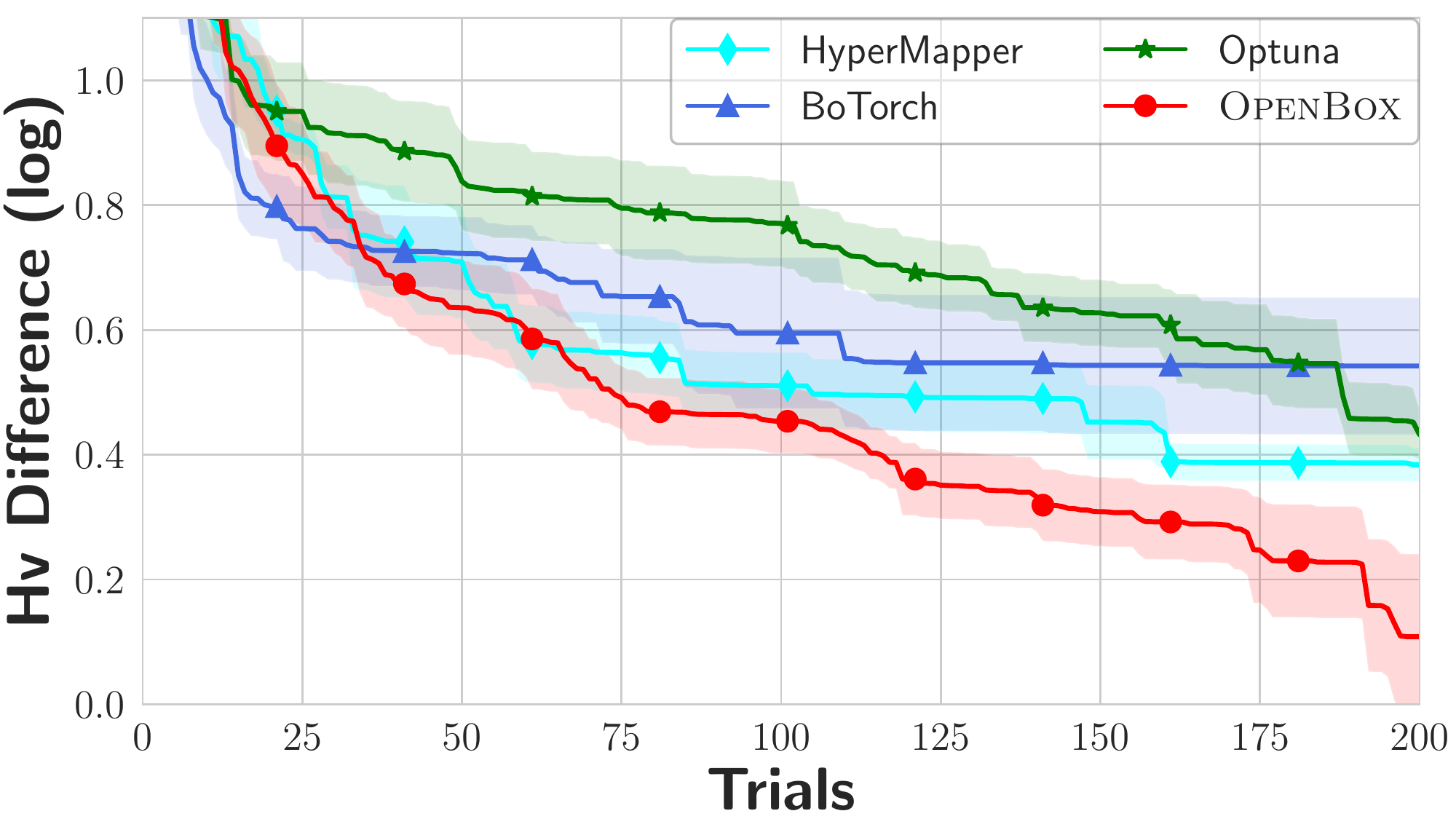}
        \label{fig:exp_constr}
}}
\subfigure[LightGBM on 24 OpenML datasets]{
    \scalebox{0.48}{
        \includegraphics[width=0.97\linewidth]{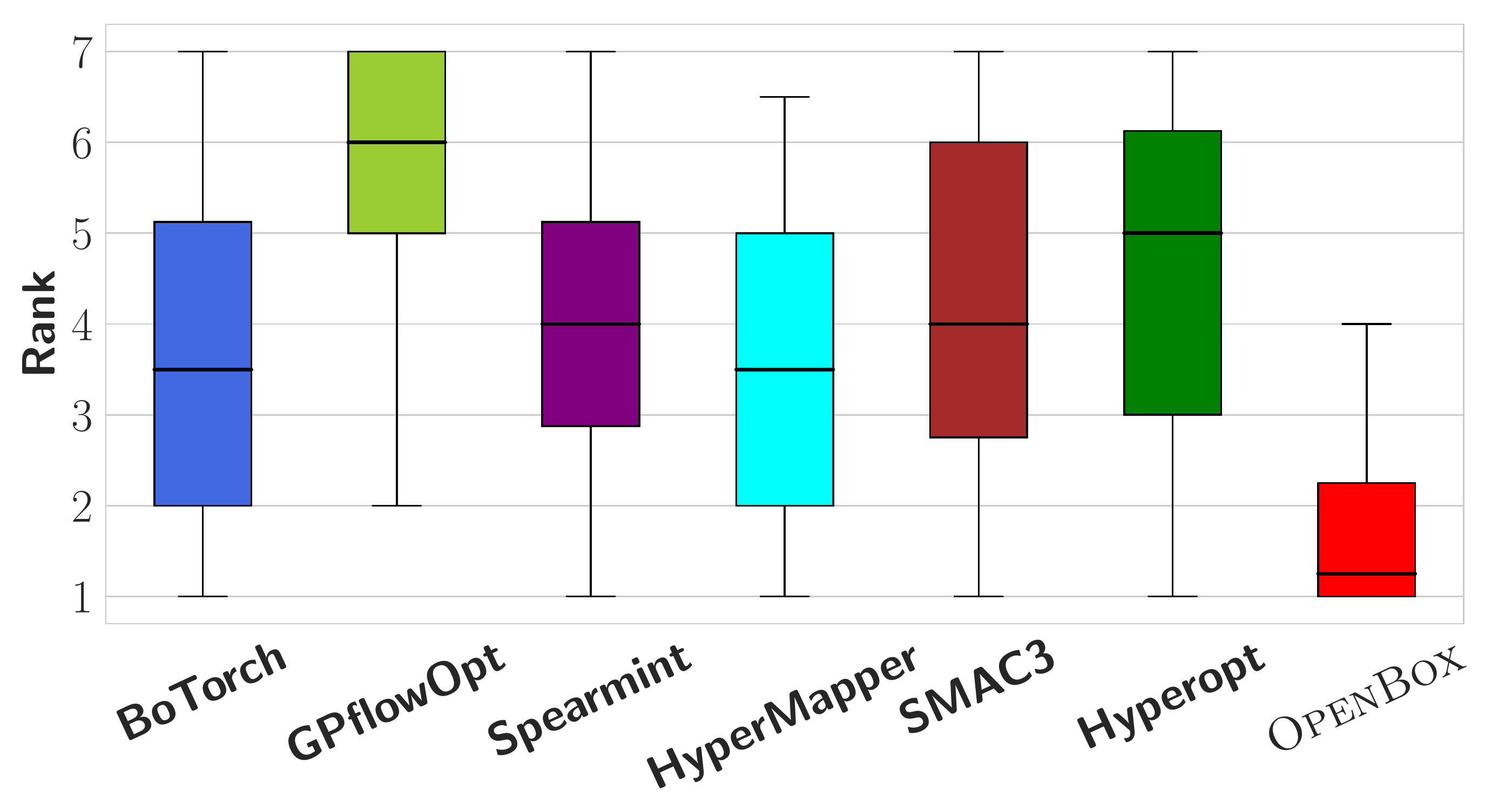}
        \label{fig:exp_lgb}
}}
\vspace{-0.5em}
\caption{Performance comparison on constrained multi-objective benchmark CONSTR (left) and LightGBM tuning task (right).}
\vspace{-1em}
\label{fig:exp}
\end{figure*}

To demonstrate the generality and efficiency of \sys, we conduct experiments
on the constrained multi-objective benchmark CONSTR and the LightGBM~\citep{ke2017lightgbm} tuning task on 24 OpenML datasets~\citep{feurer2021openml}.
We report the Hypervolume difference from the optimum in the CONSTR benchmark and the performance rank of the best-achieved accuracy in the LightGBM tuning task. 
In Figure~\ref{fig:exp_constr}, we observe that \sys outperforms the other baselines in terms of convergence speed and stability. 
In Figure~\ref{fig:exp_lgb}, we observe that \sys outperforms the other competitive systems, achieves a median rank of 1.25, and ranks first in 12 out of 24 datasets.

\presec
\section{Conclusion}
\postsec

This paper presents \sys, an open-source system for solving generalized BBO tasks. 
\sys hosts a wide range of state-of-the-art optimization algorithms and provides user-friendly interfaces along with comprehensive visualization functions.
Evaluations showcase the excellent performance of \sys over existing systems. 
The recently released version 0.8.3 has been tested on Linux, macOS, and Windows and can be installed easily via PyPI by \texttt{`pip install openbox'}. 
The source code of \sys is now available at \url{https://github.com/PKU-DAIR/open-box}.  
More detailed examples, APIs, and advanced usages can be found in our documentation\footnote{\url{https://open-box.readthedocs.io/}}.

\acks{We thank all contributors to this project.
This work is supported by the National Natural Science Foundation of China (No. U23B2048 and U22B2037), 
Beijing Municipal Science and Technology Project (No. Z231100010323002), 
and High-performance Computing Platform of Peking University.
Bin Cui is the corresponding author.
}

\vskip 0.2in
\nocite{*}
\bibliography{reference}
\end{document}